\documentclass[conference]{IEEEtran}
\IEEEoverridecommandlockouts
\usepackage{cite}
\usepackage{amsmath}
\usepackage{amssymb,amsfonts,textcomp}
\usepackage{caption}
\usepackage[para]{threeparttable}
\usepackage{tablefootnote}
\usepackage{float}
\usepackage{subcaption}
\usepackage{graphicx}
\usepackage{pdfpages}
\usepackage{ifluatex}
\ifluatex
  \usepackage{pdftexcmds}
  \makeatletter
  \let\pdfstrcmp\pdf@strcmp
  \let\pdffilemoddate\pdf@filemoddate
  \makeatother
\fi
\usepackage{svg}
\usepackage[bookmarks=true,breaklinks=true,letterpaper=true,colorlinks,citecolor=blue,linkcolor=blue,urlcolor=blue]{hyperref}

\usepackage{svg}
\svgpath{{./svg2/}}

\usepackage[T1]{fontenc}
\usepackage[utf8]{inputenc}
\usepackage{authblk}
\usepackage{lipsum}
\usepackage[american]{babel}

\author[1]{Mayank Senapati}
\author[1]{Manil Dev Gomony}
\author[1]{Sherif Eissa}
\author[2]{Charlotte Frenkel}
\author[1]{Henk Corporaal}
\affil[1]{Eindhoven University of Technology}
\affil[2]{Delft University of Technology}

\renewcommand{\baselinestretch}{0.96}
\begin{document}


\title{THOR - A Neuromorphic Processor with 7.29G $\text{TSOP}^2/\text{mm}^2\text{Js}$ Energy-Throughput Efficiency}

\maketitle

\begin{abstract}

Neuromorphic computing using biologically inspired Spiking Neural Networks (SNNs) is a promising solution to meet Energy-Throughput (ET) efficiency needed for edge computing devices. Neuromorphic hardware architectures that emulate SNNs in analog/mixed-signal domains have been proposed to achieve order-of-magnitude higher energy efficiency than all-digital architectures, however at the expense of limited scalability, susceptibility to noise, complex verification, and poor flexibility. On the other hand, state-of-the-art digital neuromorphic architectures focus either on achieving high energy efficiency (Joules/synaptic operation (SOP)) or throughput efficiency (SOPs/second/area), resulting in poor ET efficiency. In this work, we present THOR, an all-digital neuromorphic processor with a novel memory hierarchy and neuron update architecture that addresses both energy consumption and throughput bottlenecks. We implemented THOR in 28nm FDSOI CMOS technology and our post-layout results demonstrate an ET efficiency of 7.29G $\text{TSOP}^2/\text{mm}^2\text{Js}$ at 0.9V, 400 MHz, which represents a 3X improvement over state-of-the-art digital neuromorphic processors.

\end{abstract}

\section{Introduction}

Neuromorphic computing using biologically inspired Spiking Neural Networks (SNN) has arisen as a new paradigm that can accommodate energy and throughput requirements of edge AI processing~\cite{9138926}. Neuromorphic hardware aims to emulate human brain operations and offers various advantages over traditional systems, including sparse low-power computation and highly scalable parallel processing~\cite{fnins.2018.00774}. Energy efficiency (Joules/synaptic operation (SOP)) and throughput efficiency (SOPs/second/area) are the two key metrics to evaluate a Neuromorphic architecture for edge AI applications. We combine these two metrics into one single figure of merit called \textit{Energy-Throughput (ET)} efficiency in terms of $\text{GSOP}^2/\text{mm}^2\text{Js}$ to efficiently capture the trade-off between them and to fairly compare the different Neuromorphic architectures. Digital neuromorphic architectures have made considerable progress in recent years, however, little focus has been given to optimizing ET efficiency. 

\begin{figure}[htbp]
  \centering
  \includegraphics[page = 1, width=1\linewidth]{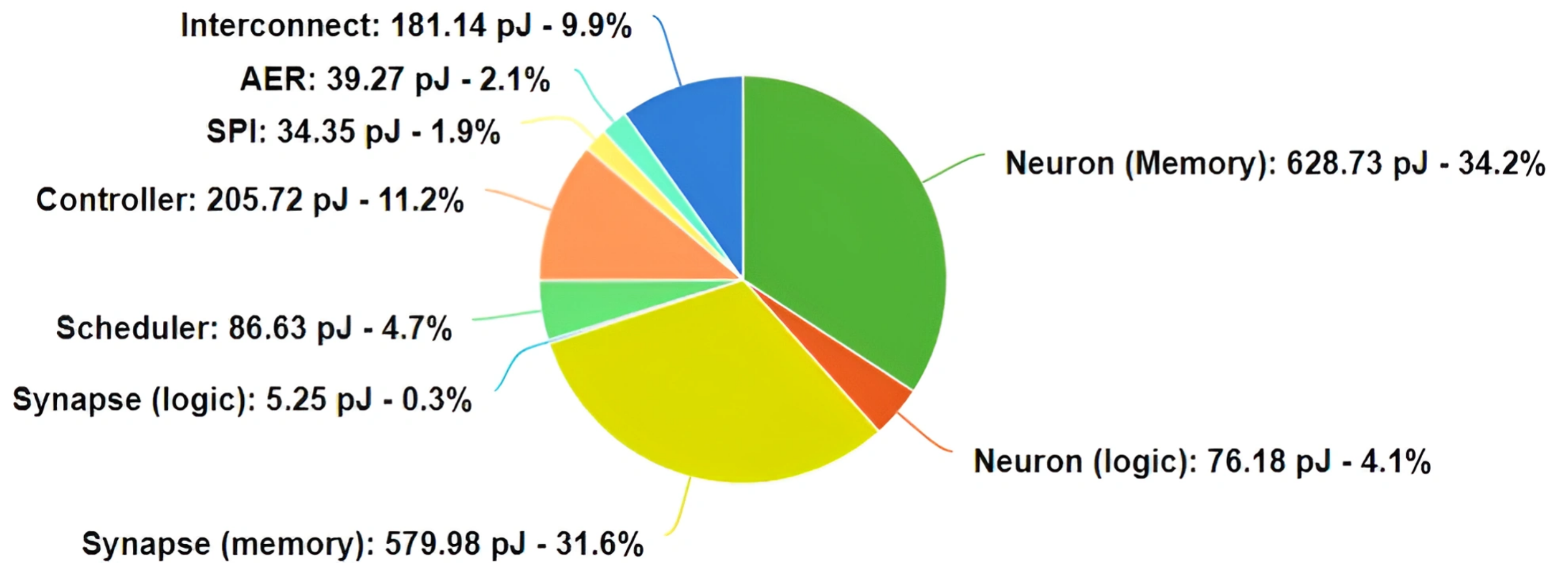}
  \caption{Energy breakdown of a state-of-the-art neuromorphic processor~\cite{frenkel20180} running MNIST for a single neuron event (using technology library 28nm FDSOI at 0.9V@100MHz). Different components of the architecture are explained in~\autoref{background}.}
  \label{odin baseline energy}
\end{figure}

Based on our analysis of energy consumption, silicon area usage and throughput of the state-of-the-art all-digital neuromorphic processors~\cite{frenkel20180, kuang202164k, wong20212, chen20184096, zhang202128nm}, we identify the following challenges that needs to be tackled for achieving the highest ET efficiency: (1) The synapse memory, which holds the individual states and parameters of the synapses is typically very large and contributes significantly to the overall energy consumption (and area usage) as can be seen in ~\autoref{odin baseline energy}, which shows the energy consumption breakdown in different components of the processor.

 SRAMs are typically used as on-chip memory in most of the all-digital neuromorphic architectures. However, SRAMs comes with different configurations in terms of number of banks, IO or word width, depth, internal multiplexing factor etc, requiring an extensive design space exploration. In addition, Standard Cell Memories (SCMs)~\cite{teman2016power} are becoming increasingly popular as a substitute to relatively smaller sized SRAMs due to high energy efficiency despite the poor area efficiency. Optimizing the energy consumption and area usage of synapse memory requires an extensive analysis of memory hierarchy for the synapse memory including different memory architectures and types. (2) The neuron and synapse memories are idle between successive accesses, which contributes to a significant amount of idle energy consumption and wastage of expensive on-chip memory bandwidth. This requires a novel architecture with efficient time multiplexing and pipelining of operations. (3) The scheduler is designed with fixed number of neurons to be processed in parallel, which is a limiting factor for scaling up the architecture for increased throughput. To address these limitations in the state-of-the-art neuromorphic processing architectures and to achieve the highest ET efficiency, this paper contributes the following:
\begin{enumerate}
    \item A neuromorphic processor THOR with a novel architecture for neuron update including a parallel neuron update scheme in the neuron event, a multi-threaded scheduler for an increased throughput, and a detailed analysis on the impact of parallelism on the energy consumption.~(\autoref{sec:architecture}) 
    \item A detailed analysis of the memory hierarchy using multiple memory types and configurations. Based on our analysis we present the  memory selection for THOR with configuration options of the different parameters (number of banks, IO or word width, depth, internal multiplexing factor etc).~(\autoref{sec:analysis}) 
    \item We perform post-layout implementation of THOR in 28nm FDSOI CMOS technology and show a high ET efficiency of 7.29 $\text{GSOP}^2/\text{mm}^2 \text{Js}$ at 0.9V, 400 MHz).~(\autoref{sec:results})   
\end{enumerate}

We review state-of-the-art architectures in \autoref{related work} and relevant background information in \autoref{background}. In \autoref{sec:architecture}, we present THOR's architecture followed by an energy exploration of different design choices and memory hierarchy in \autoref{sec:analysis} where we make our design choices. Finally, we present our implementation results in \autoref{sec:results} and make a comparison with state-of-the-art architectures and conclude our paper with \autoref{conclusions and future work}.

\section{Related work}\label{related work}

A variety of all-digital neuromorphic compute architectures have been proposed in the past. Truenorth \cite{akopyan2015truenorth}, Loihi \cite{davies2018loihi} and Spinnaker \cite{painkras2012spinnaker} represent very large scale neuromorphic architectures with multiple cores aimed at flexibility or programmability. Several multi-core architectures \cite{chen20184096, wong20212, kuang202164k, zhang202128nm}, achieve low energy consumption. For example, \cite{zhang202128nm} relies on asynchronous circuits \cite{peeters2010click} to wake memories and logic while \cite{kuang202164k} uses routing circuits that rely on spike-driven communication to keep energy consumption to a minimum. FPGA based  neuromorphic architectures \cite{huang2020spiking, mitchell2020small, irmak2021dynamic}, have been proposed as well to allow re-configurability. These architectures prioritize flexibility and programability over energy efficiency. Architectures designed for low power embedded applications \cite{lee2021neuroengine, stuijt2021mubrain, kuang202164k, zhang202128nm, wong20212, frenkel20180} achieve low energy and area usage, but run with a slow clock, achieving low throughput. For example, in~\cite{stuijt2021mubrain} event driven processing with asynchronous components acts as a bottleneck. While \cite{wang2020always} implements a time multiplexed neuron ALU with event driven clock and power gating to achieve high energy efficiency on an always-on architecture. \cite{wong20212} proposes an always-on architecture with event driven clock gating and Globally Asynchronous Locally Synchronous (GALS) architecture which achieves 2.1pJ/SOP at 0.5V .ODIN \cite{frenkel20180} uses high density SRAM memory along with a time multiplexed neuron ALU but suffers from poor ET efficiency. To summarize, state-of-the-art neuromorphic architectures with high flexibility have low energy and area efficiency while energy efficient architectures suffer from low throughput and/or area efficiency. In this paper, we aim to maximize ET efficiency of digital neuromorphic architecture by performing an extensive analysis of energy, area and throughput bottlenecks.

\section{Background}\label{background}
This section first introduces the Spiking Neural Networks (SNN) that is the class of artificial neural network supported by THOR and then the baseline hardware architecture of a digital neuromorphic processor.

\subsection{Spiking Neural Networks}
\begin{figure}[htbp]
  \centering
  \includegraphics[page = 1, width=\linewidth, trim = 0.5cm 20.5cm 1cm 1cm, clip]{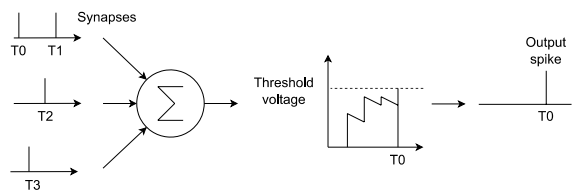} 
  \caption{Leaky Integrate and fire (LIF) neuron model. Each spike arrives at a particular synapse with a corresponding synaptic weight, which increases the membrane potential. When the membrane potential reaches a certain level, it spikes and then returns to initial value.}
  \label{lif dynamics}
\end{figure}

Spiking Neural Networks (SNN) are a class of deep learning models which attempt to mimic biological nervous systems. SNNs provide several advantages over traditional Artificial Neural Networks (ANN). The event driven nature of these networks encourages sparse computation which contributes to low power consumption. Since these networks are based on biological models, they are suitable candidates for biologically inspired online learning. The Leaky Integrate-and-fire (LIF) neuron, as shown in \autoref{lif dynamics}, is a commonly used neuron model. When a spike arrives on a synapse, it triggers an input current into the post-synaptic neuron, which is integrated as voltage called a Synaptic Operation (SOP). The voltage of the neuron leaks according to a time constant. If the neuron reaches a certain threshold voltage, it fires an output spike and resets its voltage to a resting state.

\subsection{ODIN Baseline Architecture}\label{odin intro}
\begin{figure}[htbp]
  \centering
  \includegraphics[page = 1, width=\linewidth, trim = 1cm 16.5cm 1cm 1cm, clip]{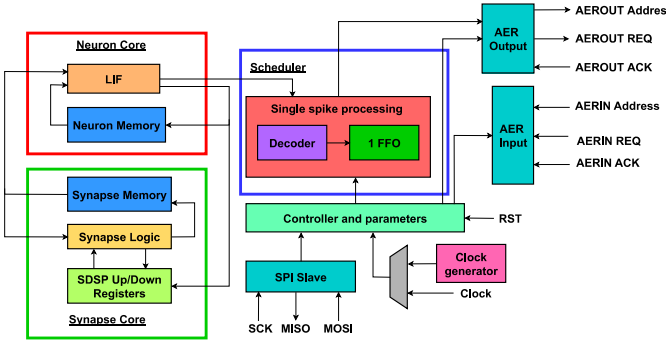}  
  \caption{Baseline ODIN Architecture after removing the logic for supporting Izhikevich neuron model consists of neuron core, synapse core and spike scheduler.}
  \label{odin baseline architecture}
\end{figure}

\begin{figure}[h]
\centering
\includegraphics[page = 1, width=\linewidth]{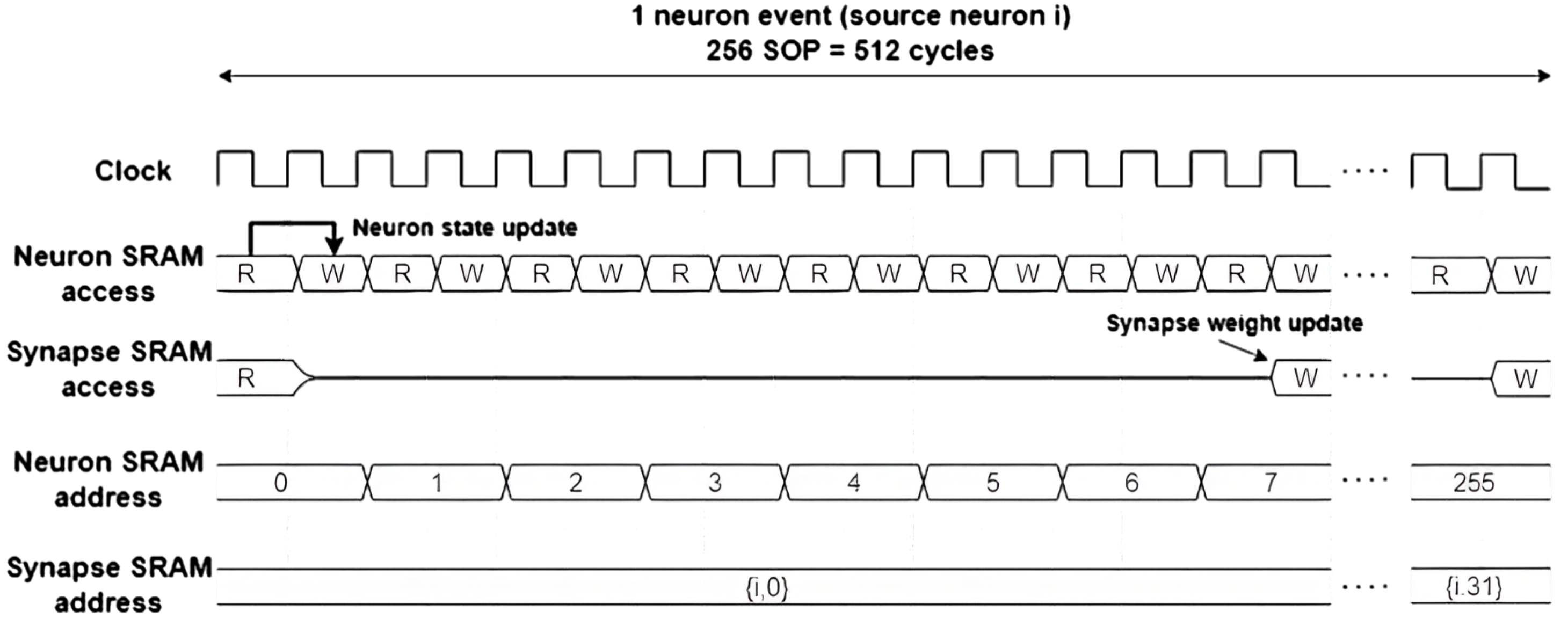}
\caption{Neuron event in ODIN showing the access patterns of neuron and synapse SRAMs\cite{frenkel20180}}
\label{ODIN neuron event}
\end{figure}

We selected the state-of-the-art digital architecture ODIN presented in~\cite{frenkel20180} as our baseline architecture, that consists of an LIF neuron core, synapse core, scheduler and peripherals as shown in \autoref{odin baseline architecture}. We reduce the original baseline by removing support for Izhikevich neurons \cite{8325231}. On-chip memories are used to store the individual states of neurons and synapses in the neuron and synapse cores, respectively. A Scheduler manages the neuron and synaptic updates. Each of the 256 neurons has a fan-in of 256 online-learning synapses, to emulated a fully connected 256x256 crossbar. An Address-event representation (AER) interface handles input and output events off-chip \cite{842110}. ODIN implements online learning with Spike-driven synaptic plasticity (SDSP) \cite{brader2007learning} and two operations: \emph{Synapse event}: Triggers one specific synaptic operation (SOP) and \emph{Neuron event}: updates all 256 neurons with 256 SOPs, according to a source neuron id, by time-multiplexing, as shown in \autoref{ODIN neuron event}. Each event takes 2 cycles per SOP and influences online learning. Each word in the synapse memory contains 8 synapses. Hence, synapse memory access takes place every 8 SOPs.

\section{THOR Architecture}\label{sec:architecture}

In this section, we first present the top-level THOR architecture and then the three main building blocks: neuron core, synapse core and multi-threaded spike scheduler with novel improvements.

\subsection{Top level architecture}\label{sec:thortoplevel}

\begin{figure}[H]
    \centering
    \includegraphics[page = 1, width=0.85\linewidth]{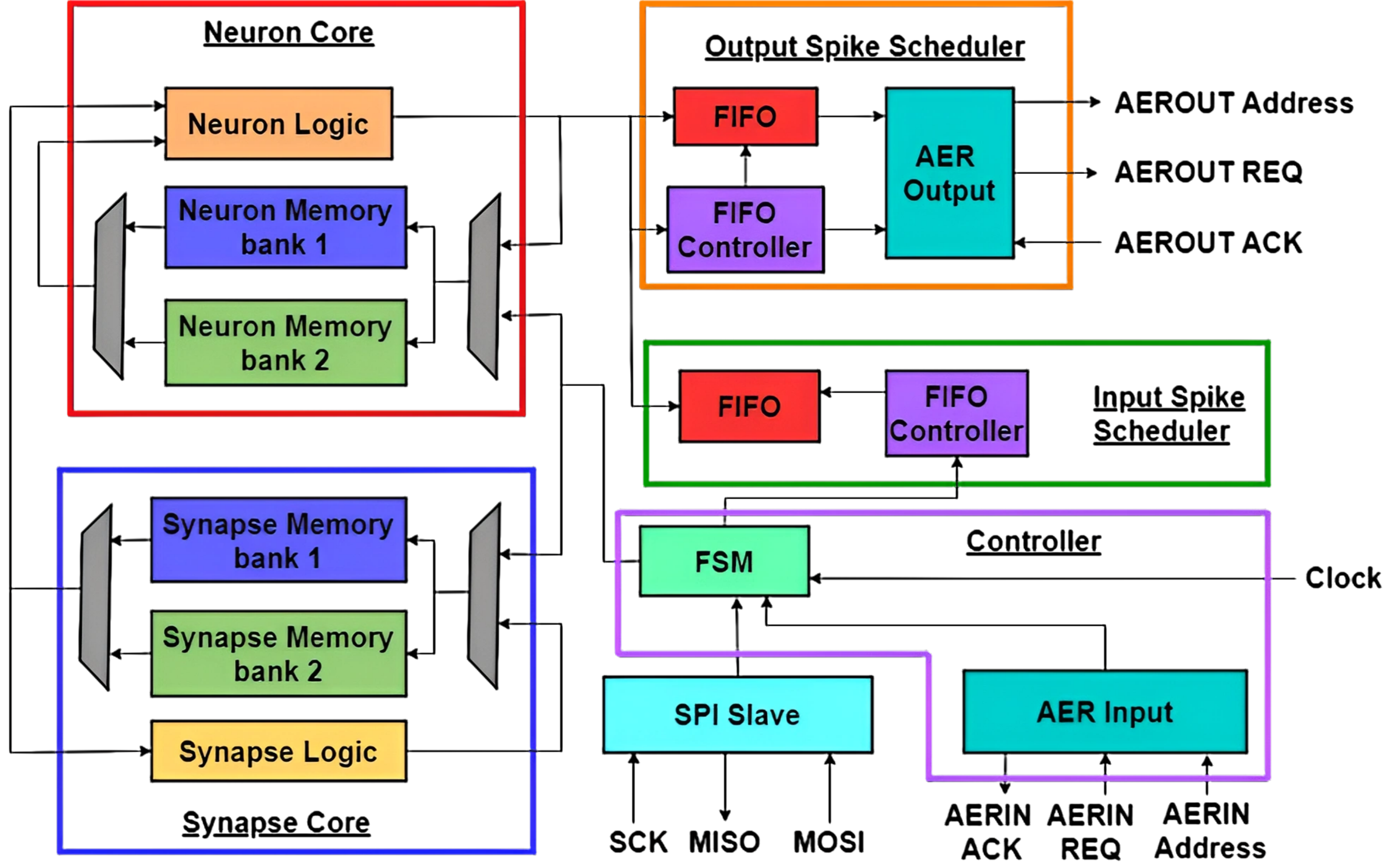}
    \caption{THOR - Top level Architecture. Detailed Scheduler architecture in \autoref{sched main}}
    \label{thor architecture}
\end{figure}

\autoref{thor architecture} shows the top level architecture of THOR. The main components are the neuron core, synapse core, and multi-threaded scheduler that consists of dedicated schedulers for output and input spikes. THOR implements an all-to-all N neuron network structure similar to the baseline ODIN architecture with the same SDSP online learning rules. Neuron and synapse memories are accessed and configured externally through the SPI interface. Input events are handled by an AER input block which is part of the controller. The input and output spike schedulers handle propagation of spikes on-chip and off-chip respectively. 

\begin{figure}[h]
\centering
\includegraphics[page = 1, width=\linewidth, trim = 0.5cm 20.5cm 1cm 1cm, clip]{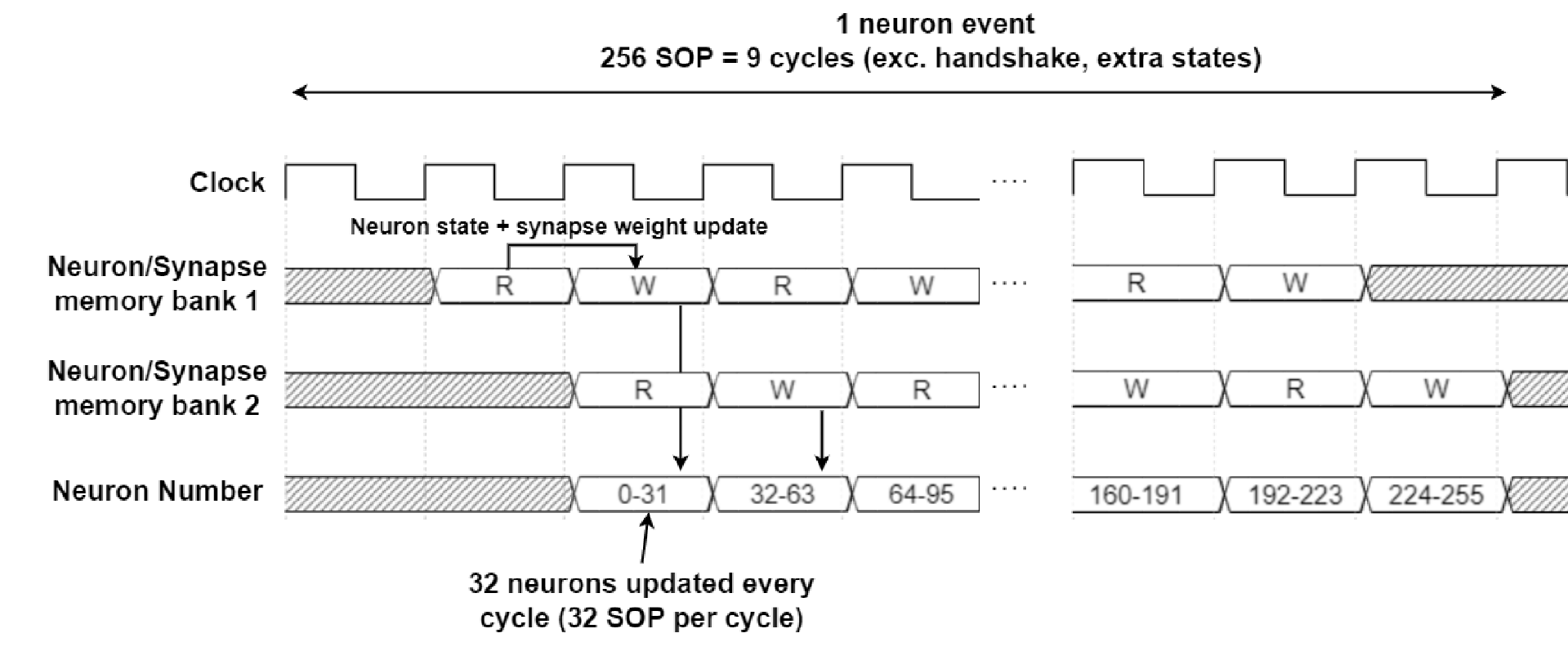}
\caption{Neuron event timing diagram of THOR processing 256 SOPs in 9 clock cycles. We have selected the configuraion of 32 neurons updates in parallel based on our analysis presented in~\autoref{sec:analysis}.}
\label{fig:neuron event timing}
\end{figure}

As the neuron event is a fundamental operation in SNNs, we propose an architecture to achieve a \emph{high-throughput neuron event} for higher area and throughput efficiency. To achieve the high-throughput neuron event, THOR contains two banks of neuron and synapse memory where each bank supports P-wide memory reads and writes. It also contains P parallel neuron and synapse logic units. Read and write operations are interleaved between the two banks to achieve high utilization of logic units. \autoref{fig:neuron event timing} shows the timing diagram of our modified neuron event where P SOPs and memory updates are executed in parallel in a pipeline. While each P SOPs take two clock cycles, we leverage our two-bank memory architecture to achieve P operation every cycle with interleaved memory access. During our modified \textit{neuron event}, all memories and logic units are fully utilized, resulting in high throughput and low leakage. 

\subsection{Neuron Core}\label{neuron core}

Our neuron core consists of two interleaved neuron memory banks and P LIF neuron logic units. The state of an LIF neuron is stored in 7 bytes as shown in \autoref{neuron state breakdown}. The LIF neuron logic consists of (1) a state update block for integration and firing. (2) a calcium update block to implement SDSP learning. Each memory bank is implemented using 7 sub-banks, one for each byte of neuron state (\autoref{neuron state breakdown}). Each sub-bank has word size of P bytes and consists of $\text{N}/2\text{P}$ entries. The leakage and threshold memory write circuits are gated during inference, as they are only configured during initialization. Furthermore, the calcium information banks can be gated and disabled if online learning is not being used.  

\begin{table}[H]
\centering
\resizebox{1\columnwidth}{!}{%
\begin{tabular}{|c|c|c|c|}
\hline
\textbf{Byte} & \textbf{Information} & \textbf{\begin{tabular}[c]{@{}c@{}}Status - \\ Online learning disabled\end{tabular}} & \textbf{\begin{tabular}[c]{@{}c@{}}Status - \\ Online learning enabled\end{tabular}} \\ \hline
0 & Membrane Potential & Read/Write & Read/Write \\ \hline
1 & Leakage & Read-only & Read-only \\ \hline
2 & Threshold & Read-only & Read-only \\ \hline
3-6 & Calcium information & Unused & \begin{tabular}[c]{@{}c@{}}Byte 4 - Read/Write\\ Byte 3,5,6 - Read-only\end{tabular} \\ \hline
\end{tabular}%
}
\caption{Neuron state breakdown (more details in \cite{frenkel20180}); 7 bytes/neuron.}
\label{neuron state breakdown}
\end{table}

\subsection{Synapse Core}

The synapse core consists of logic units and memory banks. The logic blocks are responsible for updating triggered synapses. Updates are a function of the post-synaptic membrane potential and calcium information. We implement P parallel neuron processing units in the neuron core for parallel processing.

\subsection{Multi-threaded Scheduler}\label{schedulers}
\begin{figure}[H]
\centering
\begin{subfigure}{0.5\columnwidth}
  \centering
  \includegraphics[page = 1, width=\linewidth, trim = 0.5cm 16.5cm 1cm 1cm, clip]{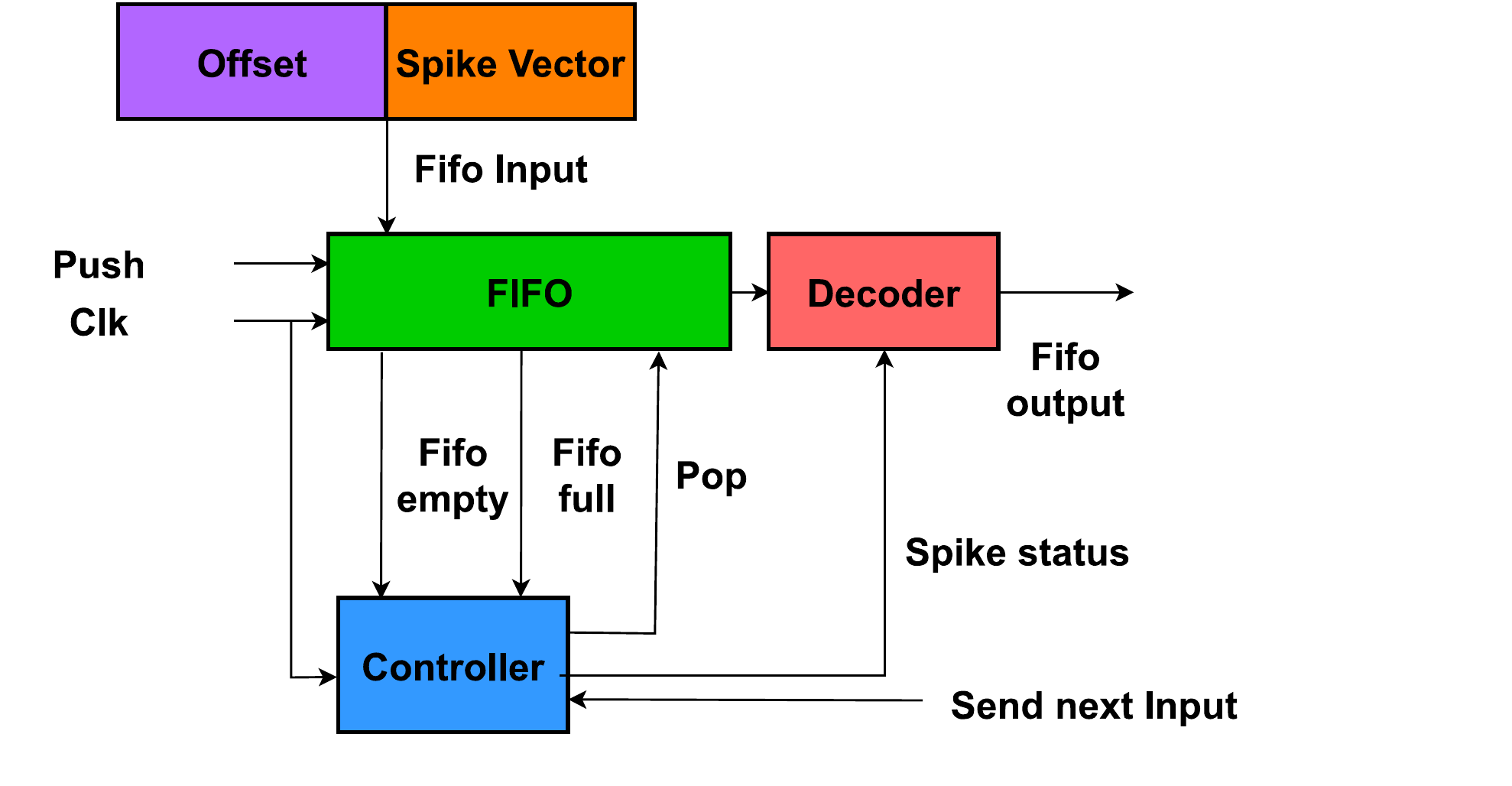} 
  \caption{}
  \label{thor sram}
\end{subfigure}%
\begin{subfigure}{.5\columnwidth}
  \centering
  \includegraphics[page = 1, width=\linewidth, trim = 0.5cm 16.5cm 1cm 1cm, clip]{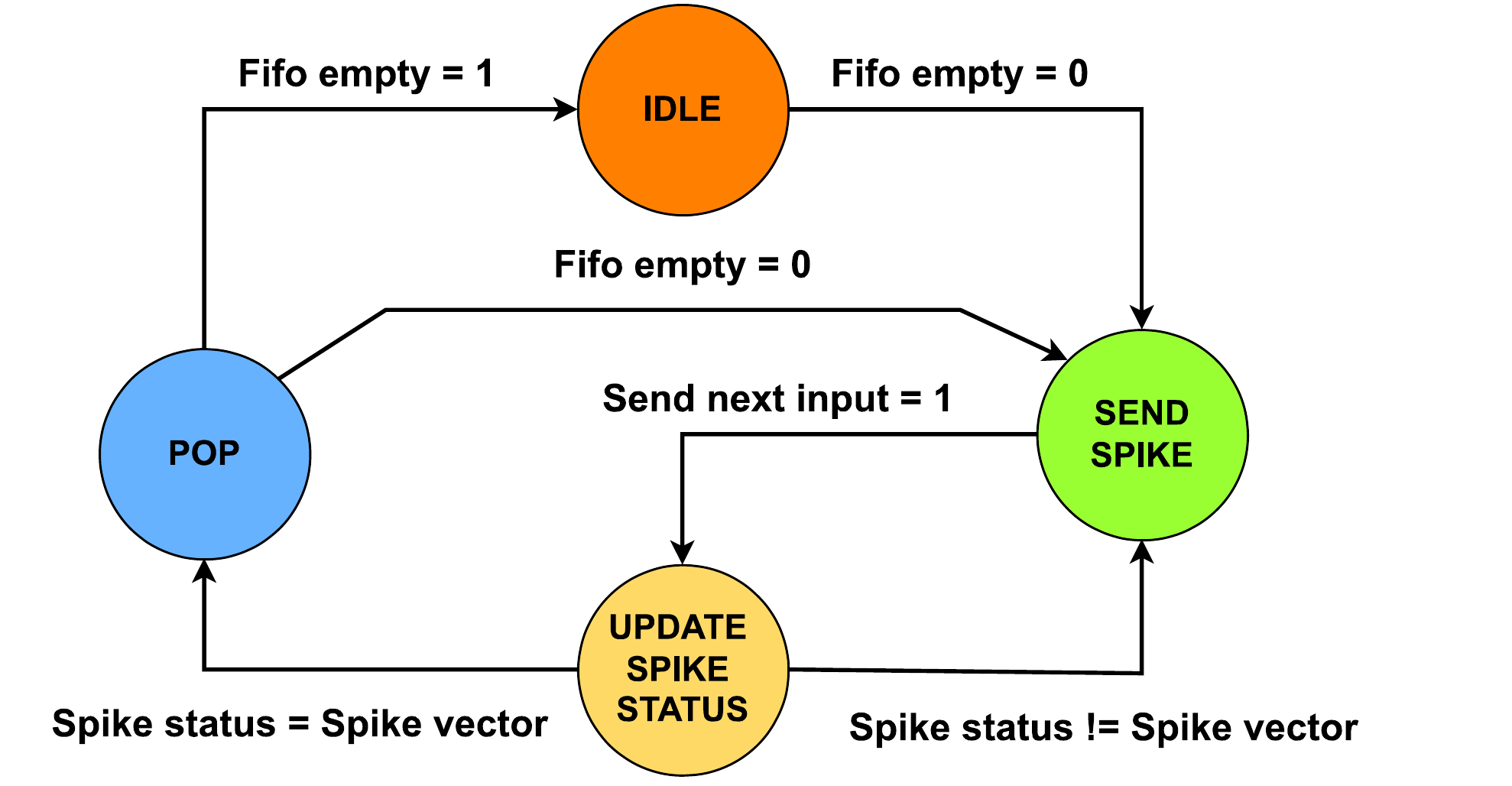} 
  \caption{}
  \label{thor scm}
\end{subfigure}
\caption{(a) Multi-threaded scheduler block diagram. The FIFO output and "Send next input" signals are connected to the AER output interface and to the controller for the output and input schedulers respectively.(b) Multi-threaded scheduler finite state machine (FSM).}
\label{sched main}
\end{figure}

We implement a multi-threaded scheduler consisting of two parallel independent hardware schedulers to handle internal and outgoing spikes. The architecture of both schedulers is identical and shown in~\autoref{sched main}~(a). They consist of a FIFO, and a controller which executes an FSM independent of the main controller. The schedulers receive a P-bit spike vector from the neuron core each cycle during a neuron event, denoting which neurons have spiked, and an offset ($log_{2} N$ bits) to indicate neuron starting address. Whenever a spike is detected in the input, the spike vector and offset are pushed into the FIFO. 

The FSM used by the controller is shown in~\autoref{sched main}~(b). While the FIFO is not empty, the scheduler decodes an output spike from the first spike vector in the queue. Output spikes are only triggered by a signal "Send next input" from the controller or the AER output, for input and output schedulers respectively, to control traffic. A dedicated status register tracks when a spike vector has been exhausted to pop it from the queue. The FIFO has a depth of $\text{N}/\text{P}$ entries, which is sufficient to handle all scenarios. Furthermore, having parallel schedulers operating independent results in higher throughput as it prevents pipeline stalls. 

\section{Memory hierarchy optimization}\label{sec:analysis}

We consider two memory types: Static Random Access Memory (SRAM) and Standard Cell Memory (SCM). SRAMs are a common choice for on chip memories and are primarily defined in terms of word size, number of words and mux factor. SCMs consist of arrays of latches or flip-flops and a readout circuit which can be built out of multiplexers, gates or tristate buffers \cite{teman2016power}. In this section, we explore the efficiency of different memory hierarchies with different parallelism schemes. We first compare SRAMs and SCMs for different bank sizes and then for different frequencies and degrees of parallelism. We conclude the section with our design choices based on analysis results.

\subsection{SCM vs SRAMs} \label{sec:SCMvsSRAM}
\begin{figure}[ht!]
  \centering
  \includegraphics[page = 1, width=\linewidth, trim = 0.5cm 18.5cm 1cm 1cm, clip]{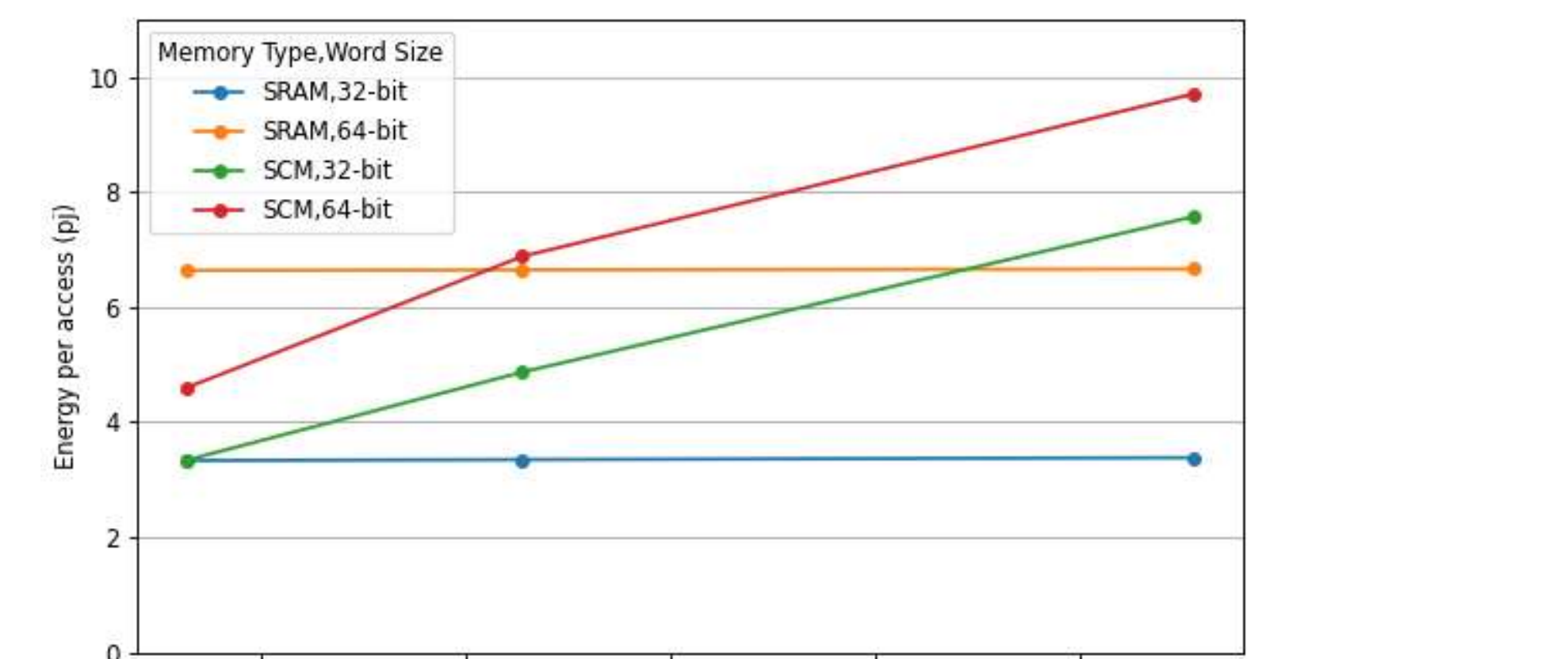} 
  \caption{Energy analysis of SCM and SRAM for different word sizes  and different memory sizes (0.9V, 100MHz) in 28nm FDSOI technology. SCM benefits from smaller memory sizes because the periphery circuitry is less complex and power hungry than that of an SRAM.}
  \label{scm sram comparision}
\end{figure}

Prior studies have shown that larger SRAMs are area efficient that SCMs, however, the latter offers better area efficiency for smaller sizes. Moreover, the energy consumption of SRAM and SCMs have not yet compared in detail for different bank sizes, multiplexing factor and I/O width. We compare the energy efficiency of SCMs and SRAMs for 32-bit and 64-bit wide word sizes and different bank sizes as shown in \autoref{scm sram comparision}. Our analysis shows that SRAM favors larger memory sizes while SCMs do not scale well with memory size due to the increasing overhead logic which increases leakage power. SCM performance improves for larger word sizes for a fixed memory size as having less entries in SCM memory reduces the complexity of the decoder and multiplexer logic. The results from \autoref{scm sram comparision} show that for specific memory and word size combinations, SCMs can have better energy efficiency than SRAMs. Furthermore, the use of low leakage libraries and optimal bank sizes improved the energy efficiency of SCMs. However, SCMs have one order-of-magnitude lower energy efficiency compared to SRAMs for large memory sizes.

\subsection{Synapse memory hierarchy} \label{sec:synmem}

\begin{figure}[h]
  \centering
   \includegraphics[page = 1, width=0.8\linewidth]{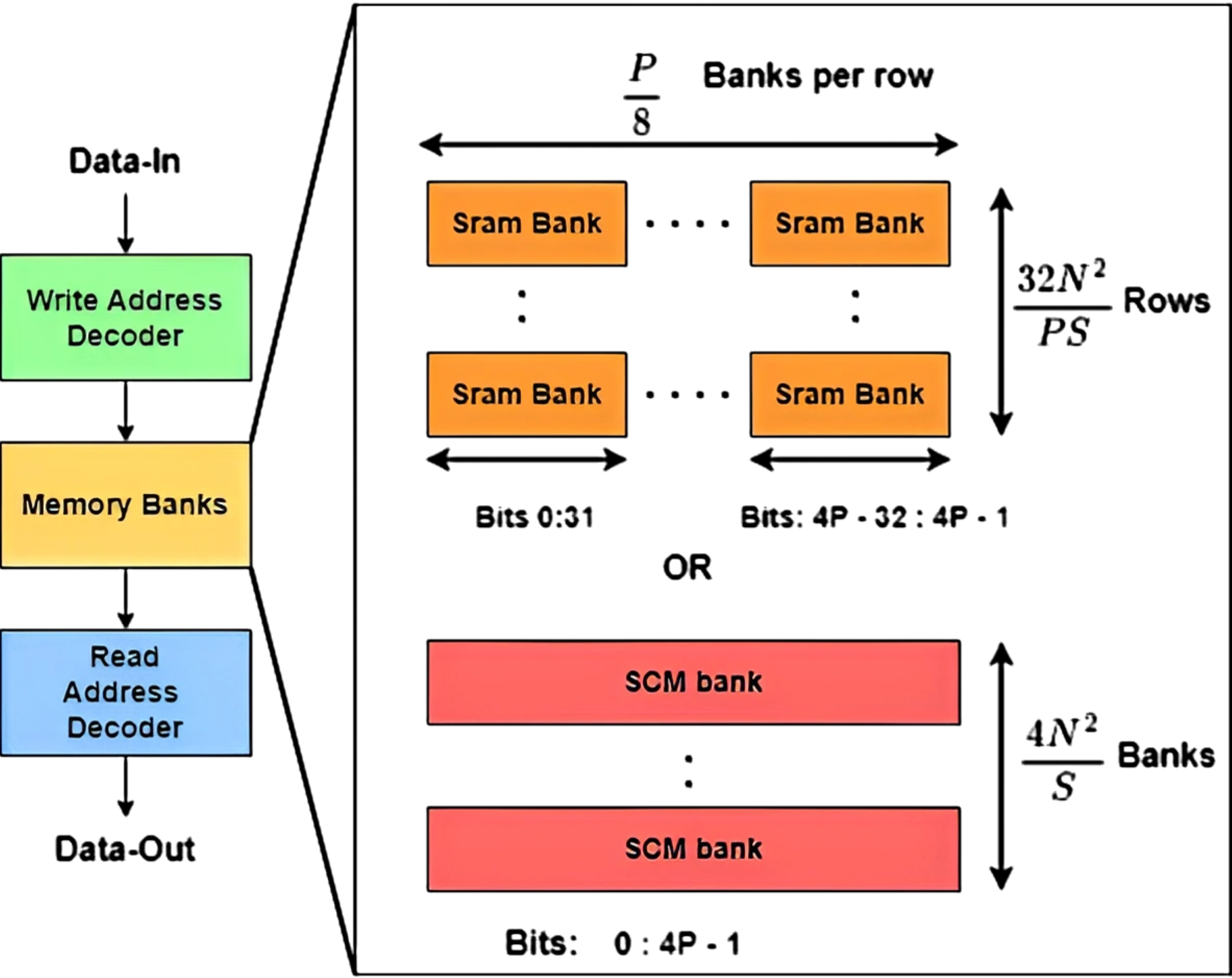}
  \caption{Synapse Memory Architecture. N = Number of neurons, P = Degree of parallelism, S = Bank size in bits}
  \label{Synapse Memory Architecture}
\end{figure}

To explore different memory hierarchies with different bank sizes and degree of parallelism, we define our synapse memory structure in a generalized manner. Let N be the number of neurons, S be the memory bank size in bits, and P be the degree of neuron update parallelism in our design. The synapse memory consists of $\text{N}^2$ synapses arranged in a crossbar architecture. The total size of the synapse memory is $4\text{N}^2$ bits and the I/O word width is $4\text{P}$ bits, as we store synapses in 4 bits (weights). \autoref{Synapse Memory Architecture} shows the general architecture of the synapse memory. For a given bank size (S), both memory types consist of $ 4\text{N}^{2}/S$ bank rows. However, as our analysis was limited by 32-bit word size SRAM macros, we have to partition each SRAM row into 32-bit wide banks ($4\text{P}/32$ banks), while for SCMs we only have 1 bank per row as we could adjust the SCM word size freely. Defining the memory hierarchy in a generalized manner allow us to create a parameterized synapse memory with SRAM macros of a specific size. The external address decoders and readout circuit are synthesized with standard cells to estimate the energy and area overhead for multiplexing. We compare SCMs and SRAMs for our energy analysis in \autoref{sec:SCMvsSRAM}. 

\subsection{Parallelism exploration}

In this section, we analyze the impact of memory choices with respect to different parallel designs for THOR. We implement designs with different amount of parallel neurons and different operating frequencies. The designs are implemented in 28nm technology with 0.9V target voltage, and synthesized with Cadence Genus. 

\begin{figure}[H]
\centering
\begin{subfigure}{0.52\columnwidth}
  \centering
  \includegraphics[page = 1, width=\linewidth]{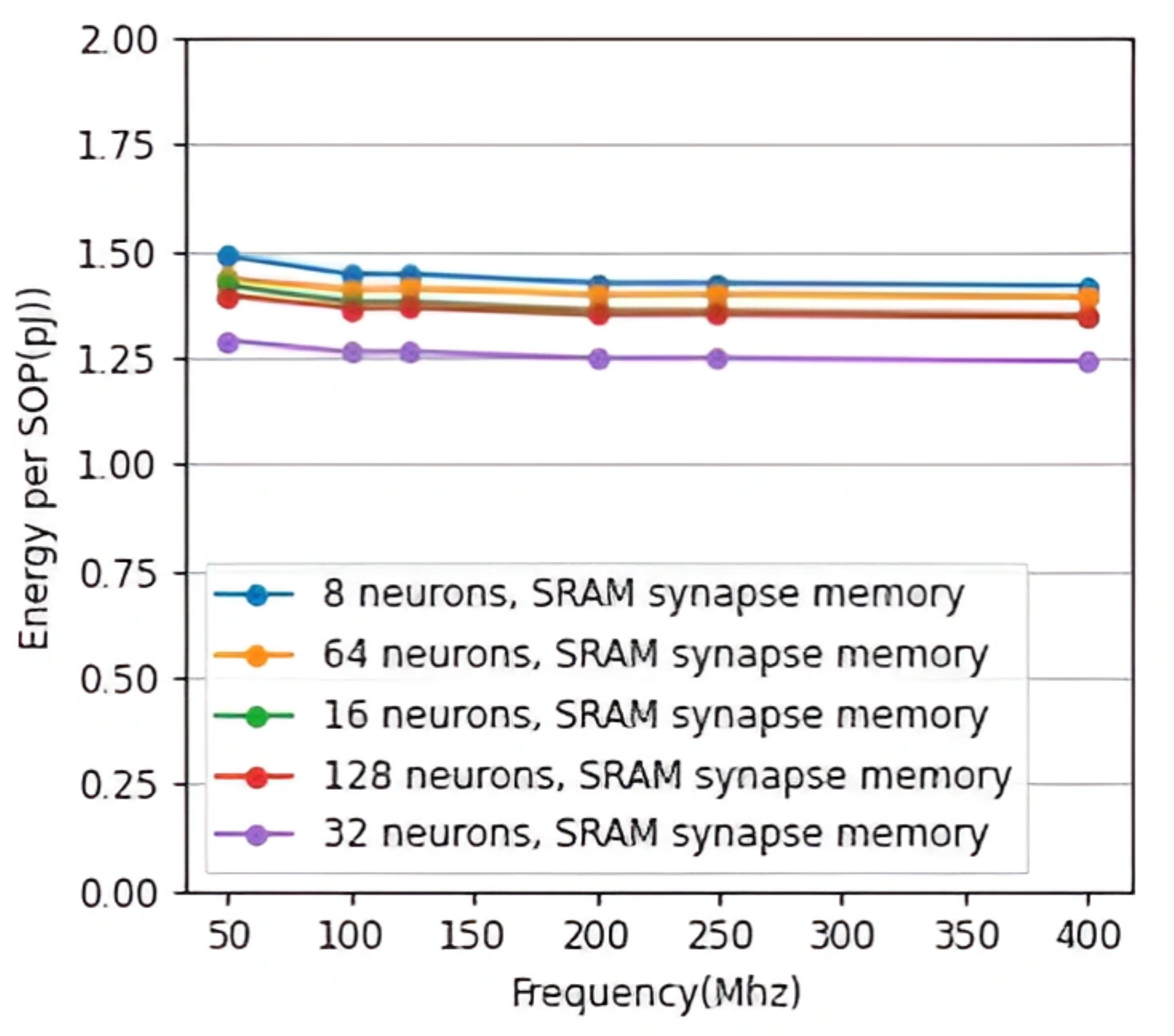}
  \caption{}
  \label{thor sram}
\end{subfigure}%
\begin{subfigure}{.5\columnwidth}
  \centering
  \includegraphics[page = 1, width=\linewidth]{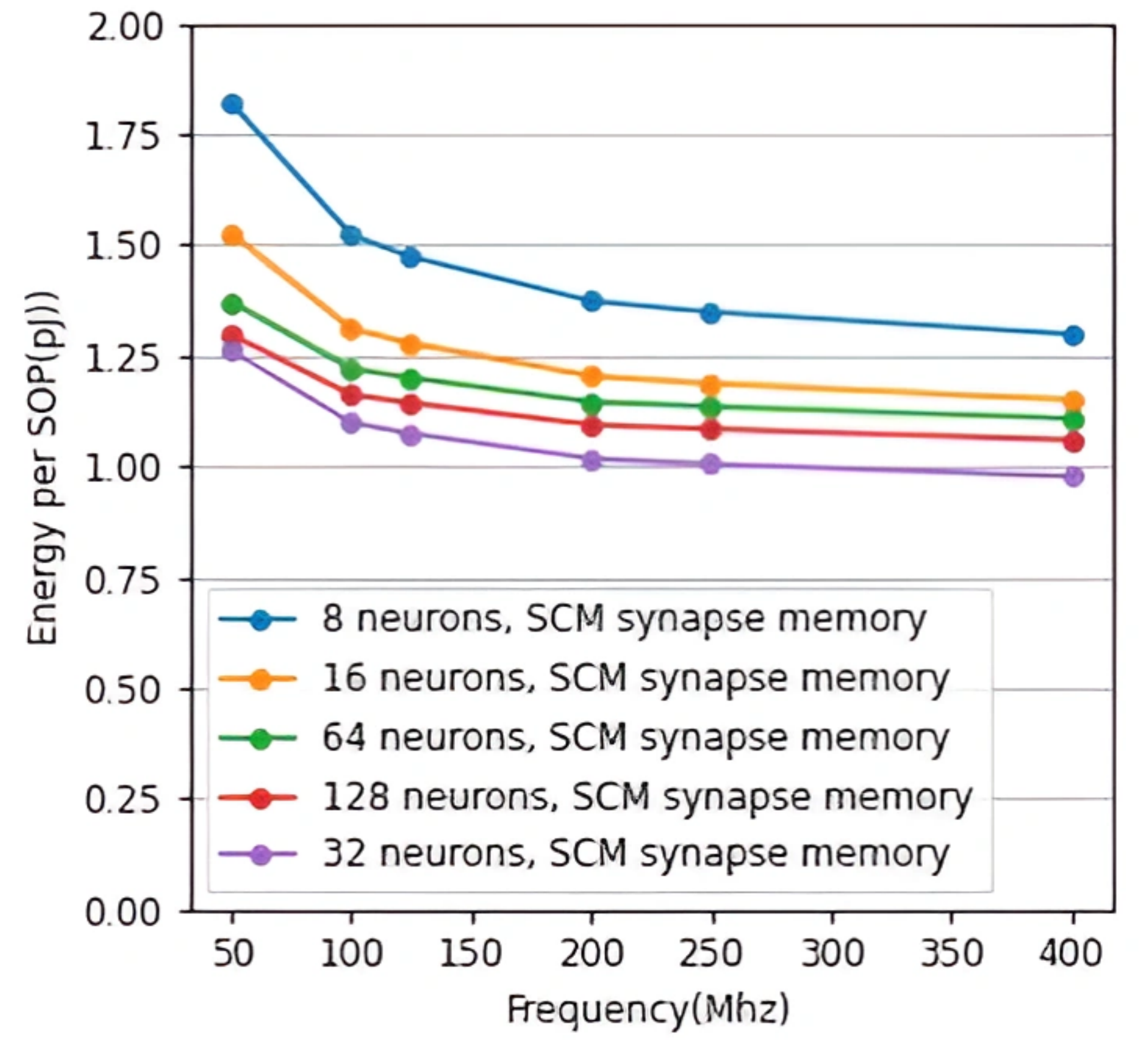}
  \caption{}
  \label{thor scm}
\end{subfigure}
\caption{$E_{sop}$ for THOR with SRAM (a) and SCM (b) (0.9V) shows that SCM based synapse memories suffer from high leakage power due to the large area of synapse memory.}
\label{thor sram scm}
\end{figure}

The power numbers are drawn from post-synthesis reports and include the overhead of AER handshakes and controller state change. The \emph{Energy per synaptic operation} $E_{sop}$ is considered as a metric of energy efficiency. We consider the scenario where the chip is saturated with neuron events. We calculate the Energy per neuron event using equation $E_{\text{SOP}} =( N_{cycles} \times T_{cycle} \times P_{avg} ) / N$, where $N_{cycles}$ is the number of cycles for a single neuron event, $N$ is the number of neurons, and $P_{avg}$ is the average power. \autoref{thor sram scm} shows the $E_{SOP}$ of different designs with different degrees of parallelism, different synapse memory technologies (synapse memory size corresponding to 64K synapses), and operating at different frequencies. At lower frequencies, SCM memories suffer from high leakage power. This can be improved using low leakage libraries. Based on the results from \autoref{thor sram scm}, we chose SCM memories with 32 paralell neuron updates for our implementation as it provides the highest energy efficiency. Note that although the energy efficiency of SRAM is very close to the SCM and yet provides a better area efficiency, we selected SCM in the final implementation to support Voltage-Frequency Scaling (which will be limited by SRAMs) in our future work.

\section{Results and Comparison}\label{sec:results}
\begin{figure}[H]
    \centering
  \includegraphics[page = 1, width=\linewidth, trim = 0.5cm 6.5cm 1cm 1cm, clip]{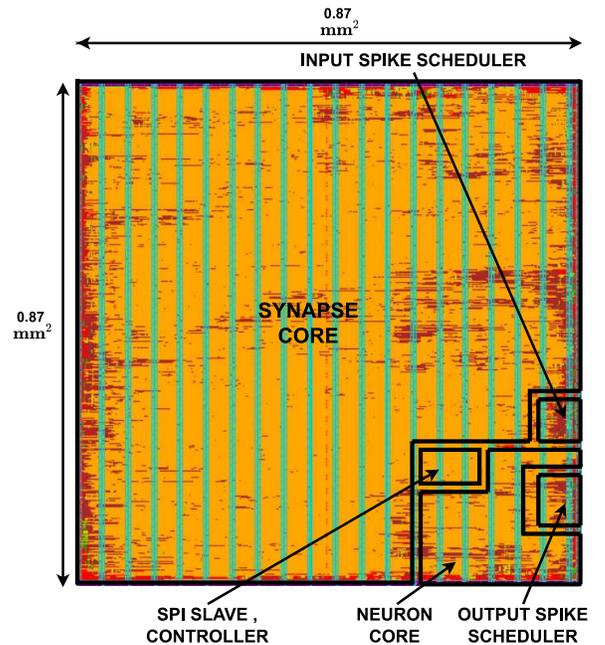} 
    \caption{THOR post-synthesis layout. Chip area ($870\mu m \times 870\mu m $) is mostly occupied by the SCM synapse memories.}
    \label{layout fig}
\end{figure}

\begin{figure}[H]
    \centering
    \includegraphics[page = 1, width=0.95\linewidth]{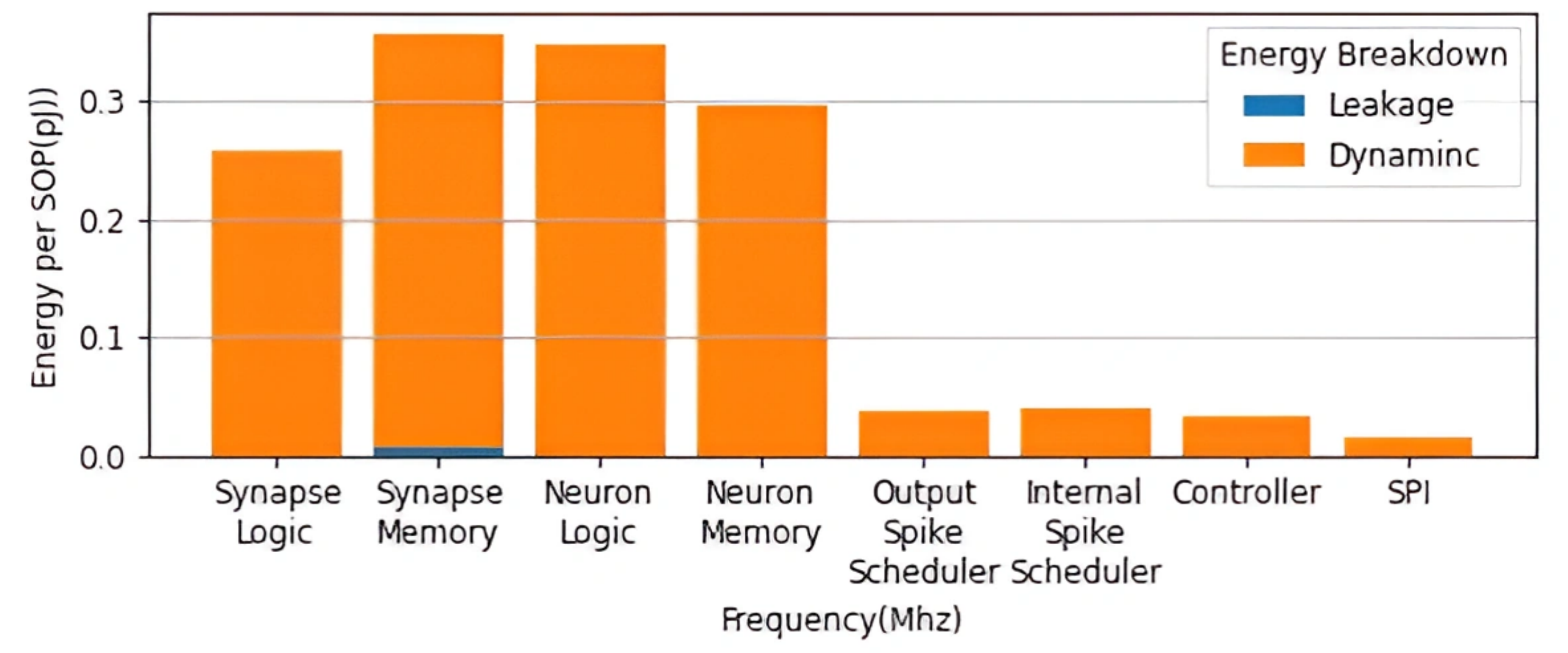}
    \caption{Post-layout Energy Breakdown of a neuron event (0.9V, 400MHz).}
    \label{layout breakdown}
\end{figure}

We implemented THOR with 32 physical neurons in 28nm FDSOI technology (\autoref{layout fig}) at 400 MHz and 0.9V, with 4KB and 64KB for the neuron and synapse memories, respectively. The post-layout netlist energy breakdown is reported in \autoref{layout breakdown}. All blocks have input and clock gating to reduce idle dynamic energy. 


As \autoref{layout breakdown} shows, the memories make significant contributions to THOR's energy consumption. We use multiple threshold voltage libraries to reduce leakage consumption. For a P-degree of parallelism, the multiplexing circuit of neuron memory must handle $7\text{P}$ bytes of neuron information. One can improve the energy efficiency further with the use of low leakage libraries and voltage-frequency scaling, especially for the synapse memories.

\begin{table*}[h]
\centering

\resizebox{18cm}{!}{%
\begin{threeparttable}[H]
\begin{tabular}{|c|c|c|c|c|c|c|c|c|}
\hline
 & \textbf{mu-brain \cite{stuijt2021mubrain}} & \textbf{Wang \cite{wang2020always}} & \textbf{Kuang \cite{kuang202164k}} & \textbf{Wong \cite{wong20212}} & \textbf{Chen \cite{chen20184096}} & \textbf{Zhang \cite{zhang202128nm}} & \textbf{ODIN}\cite{frenkel20180} & \textbf{THOR} \\ \hline
\textbf{Circuit type} & Asynchronous & Synchronous & Synchronous & Synchronous & Synchronous & Asynchronous & Synchronous & Synchronous \\ \hline
\textbf{Technology (nm)} & 40 & 65 & 65 & 40 & 10 & 28 & 28 & 28 \\ \hline
\textbf{Total area (mm2)} & 1.42 & 1.99 & 89.48 & 14.57 & 1.28 & 0.52 & 0.086 & 0.77 \\ \hline
\textbf{Number of cores} & 1 & 1 & 64 & 44 & 64 & 1 & 1 & 1 \\ \hline
\textbf{Neurons} & 336 & 650 & 64K & 11K & 64K & 256 & 256 & 256 \\ \hline
\textbf{Synapses} & 37K & 67K & 64M-total & 2.8M & 64M & 131K & 65K & 65K \\ \hline
\textbf{Online learning} & No & No & No & Yes & Yes & Yes & Yes & Yes \\ \hline
\textbf{Energy per SOP (pJ/SOP)} & \begin{tabular}[c]{@{}c@{}}0.627\\ @1.1V\end{tabular} & \begin{tabular}[c]{@{}c@{}}1.5\\ @70Khz,0.5V\end{tabular} & \begin{tabular}[c]{@{}c@{}}2.64\\ @ 24MHz,0.89V\\ 4.60\\ @192MHz,1.20V\end{tabular} & \begin{tabular}[c]{@{}c@{}}2.1\\ @12.5MHz,0.5V\\ 9.5\\ @160MHz,1V\end{tabular} & \begin{tabular}[c]{@{}c@{}}3.8 \\ @ 105MHz,0.52 V\\ 8.3 \\ @ 506MHz,0.9V\end{tabular} & \begin{tabular}[c]{@{}c@{}}3.97\\ @6.7MHz,0.8V\end{tabular} & \begin{tabular}[c]{@{}c@{}}8.40\\ @75MHz,0.55V\end{tabular} & \begin{tabular}[c]{@{}c@{}}1.40\\ @400MHz,0.9V\end{tabular} \\ \hline
\textbf{\begin{tabular}[c]{@{}c@{}}Throughput (SOP/s)\end{tabular}} & \begin{tabular}[c]{@{}c@{}}16 K\\ @ 1.4Hz \\ input rate\end{tabular} & \begin{tabular}[c]{@{}c@{}}260 K\\ @ 0.5V,70Khz\end{tabular} & \begin{tabular}[c]{@{}c@{}}12.29G\\ @ 192MHz,1,20V\end{tabular} & \begin{tabular}[c]{@{}c@{}}0.28G\\ @12.5MHz,0.5V\\ 1.93G\\ @160MHz,1V\end{tabular} & \begin{tabular}[c]{@{}c@{}}5.16G\\ @ 105MHz,0.52V\\ 25.11G\\ @ 506MHz,0.9V\end{tabular} & \begin{tabular}[c]{@{}c@{}}0.01G\\ @6.7MHz,0.8V\end{tabular} & \begin{tabular}[c]{@{}c@{}}37.5M\\ @ 75MHz,0.55V\end{tabular} & \begin{tabular}[c]{@{}c@{}}7.84G\\ @ 400MHz,0.9V\end{tabular} \\ \hline
\textbf{\begin{tabular}[c]{@{}c@{}}Energy-Throughput (ET) \\ ($\text{TSOP}^2/\text{mm}^2\text{Js}$)\end{tabular}} & \begin{tabular}[c]{@{}c@{}}17.92K\\ @1.4hz\\ input rate\end{tabular} & \begin{tabular}[c]{@{}c@{}}88.4K\\ @0.5V,70Khz\end{tabular} & \begin{tabular}[c]{@{}c@{}}0.049G\\ @ 192MHz,1,20V\end{tabular} & \begin{tabular}[c]{@{}c@{}}0.009G\\ @12.5MHz,0.5V\\ 0.014G\\ @160MHz,1V\end{tabular} & \begin{tabular}[c]{@{}c@{}}1.03G\\ @ 105MHz,0.52V\\ 2.25G\\ @ 506MHz,0.9V\end{tabular} & \begin{tabular}[c]{@{}c@{}}0.005G \\ @6.7MHz,0.8V\end{tabular} &
\begin{tabular}[c]{@{}c@{}}51.9M\\ @75MHz,0.55V\end{tabular} & \begin{tabular}[c]{@{}c@{}}\textbf{7.29G}\\ @400MHz,0.9V\end{tabular} \\ \hline
\end{tabular}

\end{threeparttable}
}
\caption{Comparison of THOR with state-of-the-art neuromorphic processors. THOR outperforms the state-of-the-art architectures by a factor of over 3X in terms of ET efficiency.}
\label{comp table}
\end{table*}

A summary of THOR's performance compared to state-of-the-art all-digital designs is shown in \autoref{comp table}. $\mu \text{Brain}$ outperforms others in $E_{sop}$ and Energy-Area efficiency, as it uses asynchronous design techniques and operates at extreme low clock frequency (in the Hz range) to reduce the dynamic power consumption. However, THOR has comparable $E_{sop}$ even with the use of synchronous logic. Moreover, THOR does uses SCM instead of SRAM to allow voltage-frequency scaling for improved energy performance. The massive parallel multi-core architectures of \cite{chen20184096} and \cite{kuang202164k} achieve high throughput, however, they have low energy efficiency. By combining all the metrics Energy consumption per SOP, Area and Throughput together, THOR outperforms state-of-the-art all-digital neuromorphic architectures by at least 3X.

\section{Conclusions}\label{conclusions and future work}
We presented THOR, an all-digital neuromorphic processor with novel architecture for neuron update including a parallel neuron update scheme in the neuron event and a multi-threaded scheduler that solves the energy and throughput bottlenecks in state-of-the-art processors. We performed an energy analysis of different memory types and configurations and devised the optimal memory hierarchy for neuron and synapse memory. We implemented THOR in 28nm FDSOI CMOS technology and demonstrated a single core THOR with an area of 0.77 $\text{mm}^2$ and an Energy-Area-Throughput efficiency of 7.37 $\text{GSOP}^2/\text{mm}^2 \text{Js}$ at 0.9V and 400 MHz, with a 3X improvement of ET efficiency compared to state-of-the-art digital neuromorphic processors.

\bibliographystyle{IEEEtran}
\bibliography{refs}

\end{document}